%% file: Formatting-Instructions-LaTeX-2026.tex
\documentclass[letterpaper]{article} 
\usepackage{aaai2026}  
\usepackage{times}  
\usepackage{helvet}  
\usepackage{courier}  
\usepackage[hyphens]{url}  
\usepackage{graphicx} 
\usepackage{natbib}  
\usepackage{caption} 
\usepackage{algorithm}
\usepackage{algorithmic}

\usepackage{enumitem}
\usepackage{subfigure}
\usepackage{multirow}
\usepackage{amsmath}
\usepackage{booktabs} 
\usepackage{amssymb}

\usepackage{newfloat}
\usepackage{listings}
\DeclareCaptionStyle{ruled}{labelfont=normalfont,labelsep=colon,strut=off} 
\floatstyle{ruled}
\newfloat{listing}{tb}{lst}{}
\floatname{listing}{Listing}
\title{Self-Correction Distillation for Structured Data Question Answering}
\author {
    Yushan Zhu\textsuperscript{\rm 1,\rm 2},
    Wen Zhang\textsuperscript{\rm 1}\thanks{Corresponding Author},
    Long Jin\textsuperscript{\rm 1},
    Mengshu Sun\textsuperscript{\rm 3}, Ling Zhong\textsuperscript{\rm 3}, Zhiqiang Liu\textsuperscript{\rm 1}, Juan Li\textsuperscript{\rm 1}, \\ Lei Liang\textsuperscript{\rm 3}, Chong Long\textsuperscript{\rm 2}, Chao Deng\textsuperscript{\rm 2}, Junlan Feng\textsuperscript{\rm 2}
}
\affiliations {
    \textsuperscript{\rm 1}Zhejiang University, China\\
    \textsuperscript{\rm 2}JIUTIAN Research, Beijing, China\\
    \textsuperscript{\rm 3}Ant Group, China\\
    zhuyushan@cmjt.chinamobile.com, \{zhang.wen, longjin\}@zju.edu.com
}

\usepackage{bibentry}

\begin{document}
\nocopyright 
\maketitle

\begin{abstract}
Structured data question answering (QA), including table QA, Knowledge Graph (KG) QA, and temporal KG QA, is a pivotal research area.        Advances in large language models (LLMs) have driven significant progress in unified structural QA frameworks like TrustUQA.     However, these frameworks face challenges when applied to small-scale LLMs since small-scale LLMs are prone to errors in generating structured queries.
To improve the structured data QA ability of small-scale LLMs, we propose a self-correction distillation (SCD) method. In SCD, an error prompt mechanism (EPM) is designed to detect errors and provide customized error messages during inference, and a two-stage distillation strategy is designed to transfer large-scale LLMs' query-generation and error-correction capabilities to small-scale LLM.
Experiments across 5 benchmarks with 3 structured data types demonstrate that our SCD achieves the best performance and superior generalization on small-scale LLM (8B) compared to other distillation methods, and closely approaches the performance of GPT4 on some datasets.  Furthermore, large-scale LLMs equipped with EPM surpass the state-of-the-art results on most datasets.
\end{abstract}


\section{Introduction}
\begin{figure}[t]
    \centering
    \includegraphics[width=1.0\linewidth]{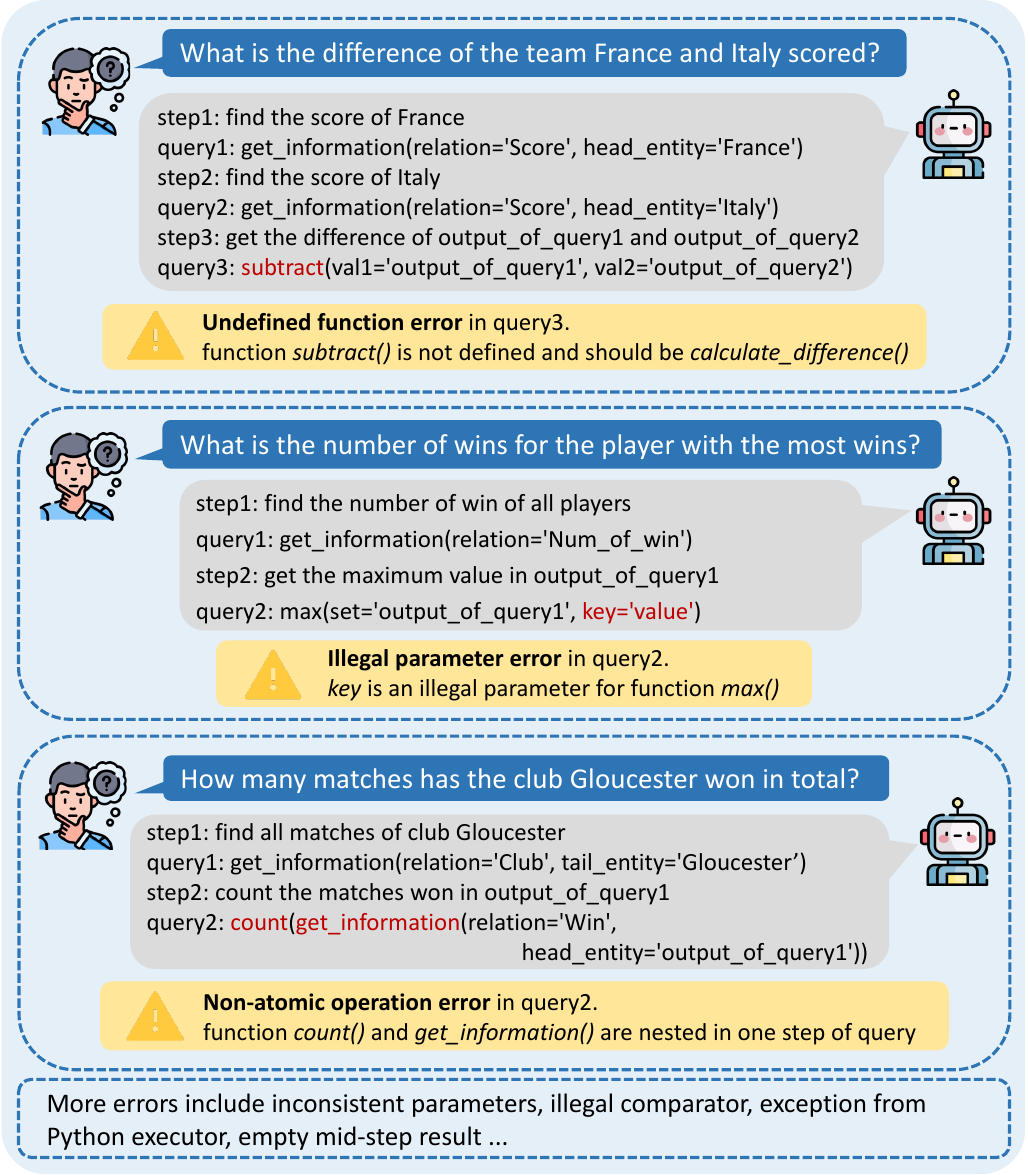}
    \caption{Examples of errors in small-scale LLM-generated queries with TrustUQA~\cite{DBLP:conf/aaai/ZhangJZ0HWHLC25}.}
    \label{fig:lm_error}
\end{figure}
Structured data question answering (QA)~\cite{DBLP:conf/aaai/ZhangJZ0HWHLC25} encompasses diverse sub-fields, including table QA~\cite{DBLP:conf/iclr/LiuCGZLCL22,DBLP:conf/emnlp/XieW0ZSYWZYWZWL22}, knowledge graph QA (KGQA)~\cite{DBLP:conf/acl/SaxenaTT20,DBLP:conf/iclr/JiangZ0W23, DBLP:conf/acl/LuoETPG0MDSLZL24} and temporal KGQA~\cite{DBLP:conf/ijcai/LiuLLGLWW0FG23,DBLP:conf/acl/ShangW0022}. 
The goal is to retrieve answers from structured data given natural language queries.
Leveraging the advanced reasoning capabilities of large language models (LLMs), recent approaches employ them as agents to generate reasoning paths \cite{DBLP:conf/iclr/SunXTW0GNSG24,DBLP:conf/iclr/LuoLHP24,xu-etal-2024-generate,DBLP:conf/acl/ChengZXYZQHCL0R24} or decompose questions into structured queries \cite{li-etal-2023-shot,DBLP:conf/acl/LuoETPG0MDSLZL24,DBLP:conf/emnlp/JiangZDYZW23,DBLP:conf/aaai/ZhangJZ0HWHLC25}.

In recent years, a growing research trend focuses on unified frameworks for multi-type structured QA, with notable advances such as StructGPT \cite{DBLP:conf/emnlp/JiangZDYZW23}, Readi \cite{DBLP:conf/acl/ChengZXYZQHCL0R24}, and TrustUQA \cite{DBLP:conf/aaai/ZhangJZ0HWHLC25}. These methods rely on API-based, closed-source LLMs (e.g., GPT3.5/GPT4) for query generation or inference.
However, in real-world applications, many users prefer independently deployed LLM solutions due to concerns over data privacy or API service stability.
Considering two practical factors in structured data QA scenarios: first, many scenarios lack the hardware capacity to deploy large-scale LLMs (over 100B parameters); second, they do not require all the general AI capabilities such as writing or solving complex math problems. 
Thus, a small-scale LLM (under 10B parameters) that supports flexible deployment and excels in structured data QA is essential.

However, directly adapting unified structured QA frameworks such as TrustUQA~\cite{DBLP:conf/aaai/ZhangJZ0HWHLC25} to small-scale LLM presents significant challenges. 
Small-scale LLM demonstrates notable limitations in structured query generation, particularly in accurately identifying and decomposing interdependent subproblems within complex questions. As illustrated in Figure \ref{fig:lm_error}, small-scale LLM exhibits various errors when processing complex queries, including: calling undefined functions, supplying illegal parameters, improperly nesting function calls within single query step, etc. 

This work introduces Self-Correction Distillation (SCD), a novel Chain of Thought (CoT) distillation framework ~\cite{DBLP:conf/acl/LiHYRC023, DBLP:conf/acl/WangWLGYR23} to transfer structured QA capabilities from a large-scale teacher LLM to a small-scale student LLM. 

First, 
we propose an Error Prompt Mechanism (EPM) that facilitates iterative error correction during the inference process. 
EPM classifies LLM-generated query errors into different types and provides type-specific feedback.
This closed-loop mechanism continuously monitors outputs, detects errors, and provides targeted corrective feedback to guide LLM through correction cycles iteratively.
Second, SCD employs a two-stage distillation strategy consisting of teacher-distillation and self-distillation stages. In the teacher-distillation stage, the student learns from both correct teacher-generated queries and CoT-based error correction trajectories through supervised fine-tuning. This stage helps transfer the query-generation and error-correction capabilities from teacher to student.
The subsequent self-distillation stage enables autonomous capability refinement, where the student model iteratively improves its performance by learning from its own outputs. This stage prevents recurring errors while enhancing student's ability to generate correct queries directly. The two-stage approach transfers teacher expertise and promotes student self-improvement, addressing both error correction and prevention in small-scale LLM.

We conduct extensive experiments on 5 benchmarks spanning 3 types of structured data, following the evaluation protocol of TrustUQA~\cite{DBLP:conf/aaai/ZhangJZ0HWHLC25}, the latest unified structured QA framework. Experimental results demonstrate that: (1) the proposed method outperforms all existing distillation approaches on small-scale LLM (8B parameters); (2) our 8B LLM achieves performance competitive with large-scale LLM (GPT4) on certain datasets; and (3) when equipped with EPM, large-scale LLM (GPT4) surpasses current state-of-the-art performance across most benchmarks.
Through comprehensive model analysis, we further validate our method's strong generalization capability to unseen data. The key contributions of this work are threefold: 
\begin{itemize}
    \item We introduce Self-Correction Distillation (SCD), a novel CoT distillation framework that improves small-scale LLMs’ structured data QA performance by transferring expertise from large-scale LLMs.
    \item In SCD, we design an Error Prompt Mechanism (EPM) to detect errors in LLM-generated queries and provide type-specific corrective feedback, along with a two-stage distillation strategy to enhance small-scale LLMs’ query generation and error correction abilities.
    \item We experimentally demonstrate that our method surpasses state-of-the-art CoT distillation approaches in enhancing small-scale LLM's structured QA capability, while exhibiting strong generalization performance on unseen dataset.
\end{itemize}

\section{Related Works}
\subsection{Structured Data QA}
Structured data question answering (QA) includes sub-fields of table QA~\cite{DBLP:conf/iclr/LiuCGZLCL22,DBLP:conf/emnlp/XieW0ZSYWZYWZWL22}, knowledge graph question answering (KGQA)~\cite{DBLP:conf/acl/SaxenaTT20,DBLP:conf/iclr/JiangZ0W23, DBLP:conf/acl/LuoETPG0MDSLZL24} and temporal KGQA~\cite{DBLP:conf/ijcai/LiuLLGLWW0FG23,DBLP:conf/acl/ShangW0022}. The goal is to seek answers to a natural language question from structured data. The field primarily employs two methodological paradigms: NL2Answer and NL2Query. NL2Answer approaches generate answers directly through language models. Some methods \cite{DBLP:conf/iclr/JiangZ0W23,DBLP:conf/emnlp/SunDZMSC18,DBLP:conf/emnlp/JiangZZLW23} train text encoders for question or reasoning representation. 
TableGPT \cite{DBLP:journals/corr/abs-2307-08674} encodes table embedding and uses it to prompt-tuning LLM.
NL2Query translates questions into formal queries, offering greater interpretability through transparent execution~\cite{DBLP:conf/aaai/ZhangJZ0HWHLC25}. Representation works include ZeroNL2SQL \cite{DBLP:journals/corr/abs-2306-08891} and DIN-SQL \cite{DBLP:conf/nips/PourrezaR23} for table QA, KB-BINDER \cite{li-etal-2023-shot} for KG QA. 
Recent advances have focused on unified frameworks for cross-domain structured QA. StructGPT  \cite{DBLP:conf/emnlp/JiangZDYZW23} leverages structured interfaces to fetch relevant knowledge, linearize and input it into LLM for direct answer generation. Readi \cite{DBLP:conf/acl/ChengZXYZQHCL0R24} generates a reasoning path for a question, edits it based on feedback, collects KG evidence, and uses LLMs to generate answers based on the evidence and question. TrustUQA \cite{DBLP:conf/aaai/ZhangJZ0HWHLC25} is the lastest unified structured data QA framework. TrustUQA adopts an LLM-friendly representation method called Condition Graph, and a two-level LLM-based querying method with dynamic demonstration retrieval.

\input{table/error_messages}

\subsection{CoT Distilling}
Knowledge distillation~\cite{DBLP:journals/corr/HintonVD15} has been widely proven to be effective in transferring knowledge from a larger model (teacher) to a smaller model (student). The main idea is to optimize the parameters of student by making its output~\cite{DBLP:conf/wsdm/ZhuZCCC0C22}, distribution~\cite{DBLP:conf/aaai/FengRL021}, hidden state~\cite{DBLP:conf/emnlp/JiaoYSJCL0L20}, or attention~\cite{DBLP:journals/corr/abs-1903-12136}, etc. close to that of teacher. 
However, the black-box nature of current mainstream large-scale LLMs (e.g., GPT3.5, GPT4) limits the application of these methods. 

Many studies perform distillation by fine-tuning smaller models directly on the output or chain of thought (CoT) of large-scale LLMs~\cite{DBLP:conf/acl/HoSY23,DBLP:conf/acl/HsiehLYNFRKLP23,DBLP:conf/icml/FuPOSK23}, referred to as CoT distillation. 
SCoTD~\cite{DBLP:conf/acl/LiHYRC023} introduces symbolic distillation to transfer CoT reasoning ability.
PERsD~\cite{chen-etal-2023-personalized} addresses the knowledge gap between student and teacher 
for code generation. Instead of fine-tuning student with correct code from teacher, PERsD enables teacher to refine the student's incorrect code. 
KPOD~\cite{DBLP:conf/icml/FengLZ00W24} imitates human cognitive patterns, training student from simple tasks and gradually to more challenging tasks. 
SCOTT~\cite{DBLP:conf/acl/WangWLGYR23} enables the student to reason faithfully through counterfactual training.
MCC-KD~\cite{DBLP:conf/emnlp/ChenWQWYZ23}  focuses on CoT reasoning diversity and consistency. 
To mitigate the impact of teacher hallucinations or errors on the student, \cite{DBLP:conf/coling/YangT0XL24} filters out the incorrect CoT and conclusions. 
OmniD~\cite{DBLP:conf/aaai/BaoSYNEK24} uses multiple sub-models to create a consensus-based pseudo-label. 
Some works~\cite{DBLP:conf/aaai/LiYFPSWW024, DBLP:conf/coling/ZhangW0024} also aim to extract useful knowledge from incorrect answers generated by teacher.

\subsection{LLM Correction}
LLM error correction aims to address reliability challenges in LLM-generated content \cite{DBLP:journals/corr/abs-2308-03188}. While early approaches like DIN-SQL \cite{DBLP:conf/nips/PourrezaR23} introduces general prompting techniques for SQL error correction, their reliance on intrinsic LLM capabilities without external validation limits effectiveness \cite{DBLP:conf/iclr/0009CMZYSZ24}. Recent work has consequently shifted toward environmental and execution feedback mechanisms, broadly categorized into prompt-based and training-based methods.

Prompt-based approaches include SPINACH's graph-exploration agent \cite{DBLP:conf/emnlp/LiuSTXZL24}, LDB's block-level verification \cite{DBLP:conf/acl/Zhong0S24}, QueryAgent's environmental feedback \cite{DBLP:conf/acl/HuangCHSXZQ24}, SALAM's dynamic demonstration retrieval \cite{DBLP:conf/emnlp/WangL23}, few-shot prompted self-correction for SQL generation \cite{DBLP:conf/iclr/ChenLSZ24}, and KB-Coder's code-style context learning \cite{DBLP:conf/aaai/NieZW024}. Although avoiding training, these methods face limited demonstration coverage and poor adaptability to small-scale LLMs with weak in-context learning.

Training-based solutions include reinforcement learning approaches like LEDEX \cite{DBLP:conf/nips/Jiang00ZHRK0D24} and PPOCoder \cite{DBLP:journals/tmlr/ShojaeeJTR23}, which suffer from reward design complexity and training instability. CoT distillation-based methods like PERsD \cite{chen-etal-2023-personalized} offer more stable training through teacher-guided correction, though their student-centric error sampling restricts exposure to diverse error types and risks data scarcity.

In this work, we reduce errors in small-scale LLM for structured QA task by synergistically combining feedback-enriched prompting with CoT distillation.

\section{Preliminary}
We briefly outline the workflow of the latest unified structured data QA method TrustUQA \cite{DBLP:conf/aaai/ZhangJZ0HWHLC25}. 
Given a question $Q$ and structured data with schema $S$, TrustUQA first converts structured data into Condition Graph (CG) format, a general and expressive data representation proposed in TrustUQA.
TrustUQA then employs a Two-layer Function-based CG Query method for querying. 
In the first layer, LLM breaks the question down into multiple query steps and write simplified functions like \textit{get\_information()} for each step, referred to as LLM queries. (The prompt template \text{P\_Q} to generate LLM queries according to the given $Q$ and $S$ is detailed in Appendix A.) The names and vocabularies of these functions (head entity, tail entity, relation, key, and value) are general and more understandable to LLM. 
In the second layer, each LLM query is translated into an execution query according to predefined rules, which are then executed over CG to retrieve the final result from structured data.

In this work, we formally define the collection of all steps and LLM queries for a question from the first layer as a ``query''. The main focus of this work is to optimize the ability of small-scale LLM to write correct ``query''. The second layer, which handles the parsing, translation, and execution of queries from the first layer, is defined as ``query executor''.

\begin{figure*}[hbpt]
  \centering
  \includegraphics[width=1.\linewidth]{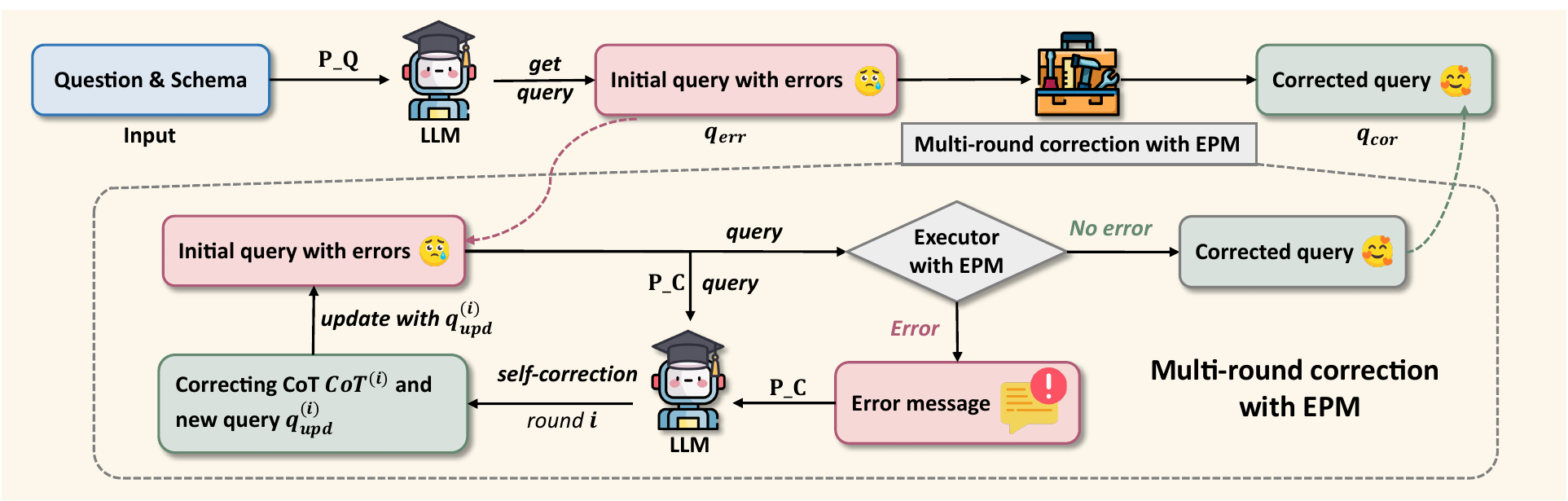}
  \caption{Multi-round Correction with EPM}
  \label{fig:reasoning}
\end{figure*}

\section{Method}
In this section, we introduce our error prompt mechanism (EPM), multi-round correction process with EPM, and two-stage distillation strategy.

\subsection{Error Prompt Mechanism (EPM)}
\label{sec:EPM}
Error prompt mechanism (EPM) is embedded in the query executor to detect errors and report customized error messages. EPM focuses on two main categories of errors in LLM-generated queries: parsing errors and execution errors. 

\subsubsection{Parsing Error Message}
Parsing error refers to the errors found during the parsing phase before query execution, including undefined function names, illegal parameters, inconsistent parameters, illegal comparators, and non-atomic operations in LLM-generated queries. The first six rows of Table \ref{tab:error_msg} are examples of each parsing error and the corresponding error message.
\subsubsection{Execution Error Message}
Execution error, usually caused by queries' logic error, is found during the execution of a query after it has successfully passed the parsing process. There are two main types: 1) exception from Python executor refers to errors that can be caught by Python's try-except statement during execution,  and 2) empty mid-step result refers to the result of a mid-step query is empty. This means that the query for that step or the previous steps may be incorrect. Examples of each execution error and the error messages are in the last two rows of Table \ref{tab:error_msg}.

Note that false negatives may occur in  the second type of execution error when the correct answer to a query is ``null/none''. In this case, the correction iteration stops after reaching the maximum correction time.

\subsubsection{Other errors}
The rest are errors unrelated to the LLM, e.g. entities or relations in the LLM-generated query are retrieved incorrectly, resulting in incorrect query results. Since such errors cannot be solved by optimizing LLM-related modules and their proportion is not high, we do not solve them in this work.

\begin{figure*}[htbp]
    \centering
    \includegraphics[width=1.\linewidth]{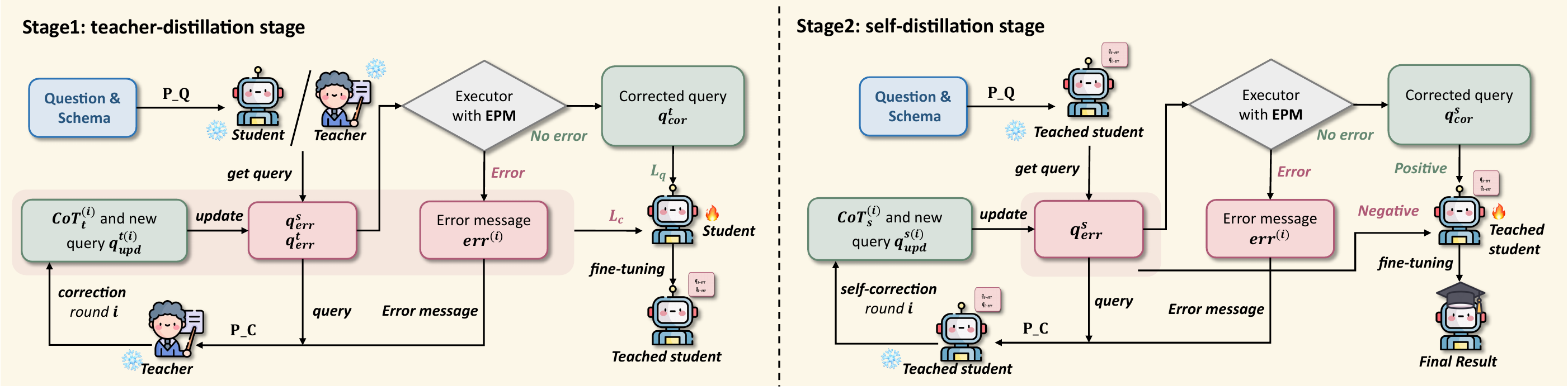}
    \caption{Two-stage Distillation Strategy}
    \label{fig:model-stage}
\end{figure*}

\subsection{Multi-round Correction with EPM}
As depicted in Figure \ref{fig:reasoning}, for a given question and data schema, LLM first writes the initial query. The query executor then parses and executes this initial query. If no error occurs, the initial query is output as the final correct query. 
Otherwise, the initial query is recorded as $q_{err}$, and the EPM reports the corresponding error message described in Section \ref{sec:EPM}. 
LLM is prompted to analyze the error's cause and correct the initial wrong query based on the error message. After updating the initial query, the query executor parses and executes it again. 
The above process iterates until no error occurs or exceeds the maximum correction time $MCT$. 
We denote the error analysis by LLMs as $CoT^{(i)}$ and the updated query as $q_{upd}^{(i)}$ after the $i$-th correction round. Assuming that after $n$ ($n \leqslant MCT$) rounds, the query $q_{upd}^{(n)}$ can be parsed and executed without error, we output it as the final correct query $q_{cor}$.
\subsection{Two-stage Distillation Strategy}
Two-stage distillation strategy includes \textit{teacher-distillation} and \textit{self-distillation} stages, where large-scale LLM serves as teacher and small-scale LLM serves as student.

\subsubsection{The Teacher-distillation Stage}
In the first stage, the student learns query generation and error correction capabilities from teacher. The teacher generates data for supervised fine-tuning the student. 

In left part of Figure \ref{fig:model-stage}, firstly, to fully explore and utilize the errors of different models, both student and teacher generate initial queries for a given question and data schema. The initial query with errors from student and teacher are denoted as $q_{err}^s$ and $q_{err}^t$, and the error message reported by EPM is denoted as $err^{(0)}$. 
Then $q_{err}^s$ or $q_{err}^t$ and $err^{(0)}$ are input into teacher for multiple rounds of error correction. In the $i$-th correction round, the error analysis by teacher is denoted as $CoT_t^{(i)}$, and the updated query is denoted as $q_{upd}^{t(i)}$. After $n$ rounds of correction, if the query $q_{upd}^{t(n)}$ can be executed without errors and yields the correct answer, it is denoted as $q_{cor}^t$. Note that $q_{cor}^t$ represents all correct queries from the teacher, both those generated directly by teacher and those corrected from teacher-generated or student-generated wrong queries.

To train the student's query generation ability, question $Q$ and data schema $S$ are input into student, and $q_{cor}^t$ is the target output:
\begin{equation}
\mathcal{L}_{q} = -\sum_{j}\text{log}P_{\mathcal{M}}\big(q_{cor(j)}^t | \text{P\_Q}(Q, S);q_{cor(<j)}^{t}\big)
    \label{equ:L-gen_query}
\end{equation}
where $q_{cor(j)}^t$ is the $j$-th token of $q_{cor}^t$, $\text{P\_Q}$ is the prompt template~\cite{DBLP:conf/aaai/ZhangJZ0HWHLC25} of query generation, and more details are in Appendix A. To train the student's error correction ability, for each correction round $i$, question $Q$, data schema $S$, wrong query $q_{upd}^{t(i-1)}$ from the last round, and the error message $err^{(i-1)}$ are input into student, and error analysis $CoT_t^{(i)}$ from teacher and the final correct query $q_{cor}^t$ are the target output:
\begin{equation}
\begin{aligned}
\mathcal{L}_{c} = & -\sum_{i=1}^{n}\sum_{j}\text{log}P_{\mathcal{M}}\big([CoT_{t}^{(i)};q_{cor}^t]_{(j)} \mid \\
&\text{P\_C}(Q,S,q_{upd}^{t(i-1)},err^{(i-1)});[CoT_{t}^{(i)};q_{cor}^t]_{(<j)}\big)    
\end{aligned}
    \label{equ:L-correct}
\end{equation}
where $q_{upd}^{t(0)}$ = $q_{err}^t$ or $q_{err}^s$, $[x;y]$ represents text concatenation of $x$ and $y$, $\text{P\_C}$ is the prompt template of error correction, and more details are in Appendix A.

Note that for each correction round $i$, we use the final correct $q_{cor}^{t}$ as the target output for student, rather than the updated query $q_{upd}^{t(i)}$ of that round. This is because non-final-round queries still contain errors, and learning to generate them does not contribute to the LLM's QA ability. Moreover, this design makes different correction rounds become tasks of varying difficulty: In later rounds, the degree of match between $CoT_t^{(i)}$ and $q_{cor}^t$ is higher, allowing student to generate the correct query more easily based on $CoT_t^{(i)}$. In earlier rounds, $CoT_t^{(i)}$ and $q_{cor}^t$ match less, and $CoT_t^{(i)}$ only describes part of the clues from $q_{upd}^{t(i-1)}$ to $q_{cor}^t$. So besides $CoT_t^{(i)}$, student must also fully understand the association between data schema and problem decomposition, which is more challenging.

The final loss of the first distillation stage is
\begin{equation}
    \mathcal{L}_{1} = \mathcal{L}_q + \mathcal{L}_c
    \label{equ:stage1}
\end{equation}

\subsubsection{The Self-distillation Stage}
After the first stage, student acquires basic query generation and error correction capabilities. In this stage, student improves itself based on its current output instead of learning from teacher, aiming to write correct queries with minimal corrections.

In right part of Figure \ref{fig:model-stage}, for a given question and data schema, student writes the initial query, denoted as $q_{err}^s$ if an error occurs in the query executor, with error message $err^{(0)}$. 
Then the wrong query $q_{err}^s$ and error message $err^{(0)}$ are feedback to itself for multiple rounds of error correction. 
The updated query after the $i$-th correction round is $q_{upd}^{s(i)}$. 
After $n$ rounds, query $q_{upd}^{s(n)}$ is denoted as $q_{cor}^s$ if it can be executed with no error and yields the correct answer. 
Since $q_{upd}^{s(i)}$ $(0\leqslant i<n)$ is the wrong query that current student more likely to generate, for the given question and data schema, we increase the generation probability of $q_{cor}^s$ and reduce the generation probability of $q_{upd}^{s(i)}$ $(0\leqslant i<n)$. The loss of the second distillation stage is
\begin{equation}
\begin{aligned}
& \mathcal{S}(q)=\sum_{j=1}^{|q|} \log P_{\mathcal{M}}\left(q_{(j)} |\text{P\_Q}(Q, S);q_{(<j)}\right) \\
& \mathcal{L}_2=-\sum_{i=1}^n\left(S\left(q_{c o r}^s\right)-S(q_{u p d}^{s(i-1)})\right),
\end{aligned}
\end{equation}
where $q_{upd}^{s(0)}$ = $q_{err}^s$.  

\section{Experiments}

\subsection{Dataset}
Following~\cite{DBLP:conf/aaai/ZhangJZ0HWHLC25}, we adopt 3 types of structured data: table with WikiSQL~\cite{DBLP:journals/corr/abs-1709-00103} and WTQ~\cite{DBLP:conf/acl/PasupatL15}, KG with WebQuestionsSP(WebQSP)~\cite{DBLP:conf/acl/YihRMCS16} and MetaQA~\cite{DBLP:conf/aaai/ZhangDKSS18}, and temporal KG with CronQuestions~\cite{DBLP:conf/acl/SaxenaCT20}.  

\input{table/datasets}
\input{table/main_result}

Due to the unbalanced distribution of datasets and the high cost of using OpenAI, we employ uniform sampling for excessively large data to construct our final training dataset. For MetaQA, we sample 30,000 training and 6,000 development instances according to each hop's proportion. For CronQuestion, we sample  25,000 training instances and 4,000 development instances according to each class's proportion. For WikiSQL, we sample 3,000 development instances. Since WTQ has no validation set, we split the raw training set of WTQ into random 80-20 splits for development following~\cite{DBLP:conf/acl/PasupatL15}. The statistical details of the original dataset (\#Raw) and our sampled dataset for training (\#SFT) are presented in Table \ref{tab:datasets}.

\subsection{Baseline}
We choose GPT4 as teacher and Llama3.1-8B-Instruct~\cite{llama3.1,DBLP:journals/corr/abs-2407-21783} as student. 

As our method is fundamentally a CoT distillation approach, we compare against representative CoT distillation baselines without distinguishing between code or text generation tasks. The baseline methods include: 
\textbf{(1) Llama3.1}, the original Llama3.1 base model;
\textbf{(2) Naive-SFT}, the most commonly used SFT strategy, which trains on error-free executable queries (both directly generated and corrected by the teacher) but neglects error correction capability;
\textbf{(3) PERsD}~\cite{chen-etal-2023-personalized}, a code correction CoT distillation method, focuses on generating personalized data by teacher refining student output, with training data that excludes teacher-generated correct queries and error correction trajectories of teacher-generated wrong queries; 
\textbf{(4) KPOD}~\cite{DBLP:conf/icml/FengLZ00W24}, the state-of-the-art CoT distillation method, employing progressive task difficult by decomposing complete queries into step-wise subtasks, ranging from generating subsequent steps given preceding ones (simple) to full query generation (complex). 

We evaluate two inference configurations: \textbf{(1) inference w/ EPM}, where LLMs perform iterative error correction during multi-round reasoning, and \textbf{(2) inference w/o EPM}, limited to single-pass reasoning.

\subsection{Experiment Setting}
We implement our method by LLama-Factory~\cite{zheng2024llamafactory}, an open-source LLM fine-tuning framework based on PyTorch. We use the LoRA~\cite{DBLP:conf/iclr/HuSWALWWC22} fine-tuning strategy and train the student on 2*NVIDIA A100 PCIe 40GB GPUs with per device batch size as $1$ and gradient accumulation steps as $8$ for $3$ epochs for each distillation stage. We use fp16 training with AdamW~\cite{DBLP:conf/iclr/LoshchilovH19} optimizer whose learning rate is $0.0001$, together with a cosine learning rate scheduler with a warm-up ratio of $0.1$. Same as previous work~\cite{DBLP:conf/aaai/ZhangJZ0HWHLC25}, we use SentenceBERT~\cite{DBLP:conf/emnlp/ReimersG19} as the dense text encoder, set the number of retrieves to 15, the number of query generation demonstrations in \text{P\_Q} to 8 and use self-consistency strategy of 5 times. For multi-round error correction, we set the number of error correction demonstrations in \text{P\_C} to 8 and the max correction time to 3. Detailed specifications of EPM implementation and error type provenance are in Appendix B.

\subsection{Result Analysis}
Table \ref{tab:main_result} shows the results on different datasets.
Teacher ``inference w/ EPM'' improves significantly than ``inference w/o EPM'' and outperforms SOTA on most datasets, demonstrating the effectiveness of EPM.
Although our student model does not achieve SOTA, it outperforms other distillation baseline methods on almost all datasets.
The results of the raw Llama3.1 are significantly worse than others. 
For Llama3.1 and Naive-SFT, ``inference w/ EPM'' shows limited improvement compared to ``inference w/o EPM'', indicating that small-scale LLMs can't obtain good error correction capability without specific fine-tuning.

\input{table/ablation_detail}
\subsection{Model analysis}
\subsubsection{Correction of Different Error Types}
\begin{figure}[hbpt]
    \centering
    \includegraphics[width=1.0\linewidth]{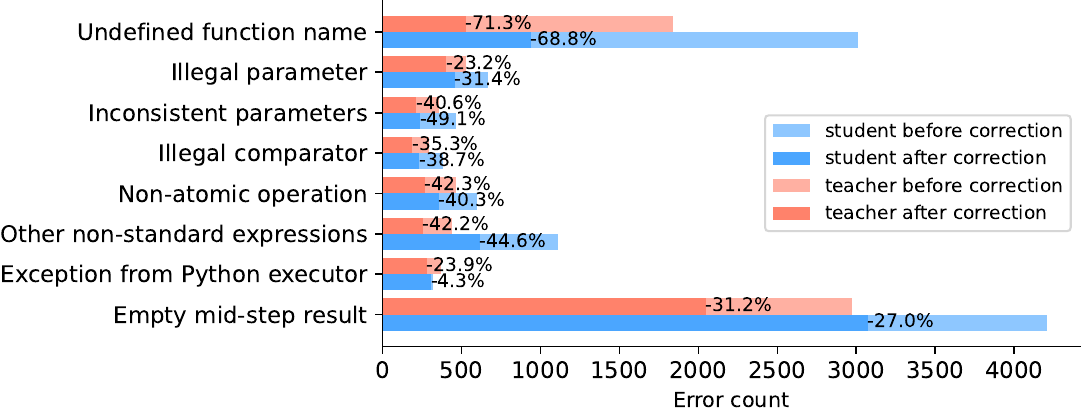}
    \caption{Error count before/after correction on all test data. Bar values are the percentage of errors corrected.}
    \label{fig:error_EPM}
\end{figure}
Figure \ref{fig:error_EPM} shows the most common error type is the empty mid-step result of execution errors, followed by undefined function names in parsing errors. The average correction rate of parsing errors reaches $40.2\%$ and $45.5\%$ for teacher and student, while on execution errors it's only $27.6\%$ and $15.7\%$ for teacher and student. 
Overall, parsing errors are easier to correct than execution errors, likely because parsing errors are often superficial and identifiable, whereas the root causes of execution errors are harder to pinpoint.

\subsubsection{Ablation Study}
Table \ref{tab:ablation} shows that without the second self-distillation stage (\textit{- self-distillation}) or without specific error messages during the multiple rounds of error correction (\textit{- error-msg in EPM}), the performance of the student model significantly declines in all datasets. Especially on more complex problems such as 3-hops of MetaQA and complex class of Cronquestion, the performance is notably diminished if LLMs do not obtain the error details during error correction.

\subsubsection{Hyperparameter Analysis}
\begin{figure}[h]
    \centering
    \subfigure[Demonstration Size]{\includegraphics[width=0.456\linewidth]{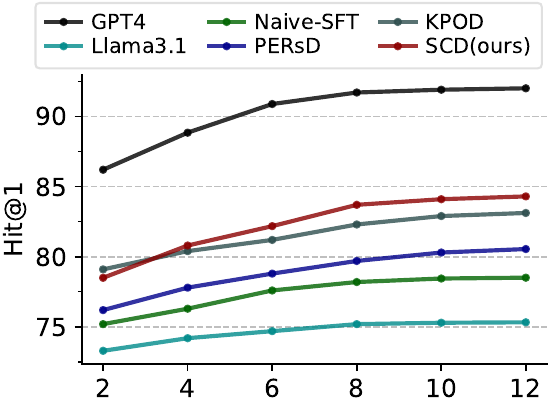}
    \label{fig:hyper_demo}}
    \subfigure[Max Correction Time]{\includegraphics[width=0.456\linewidth]{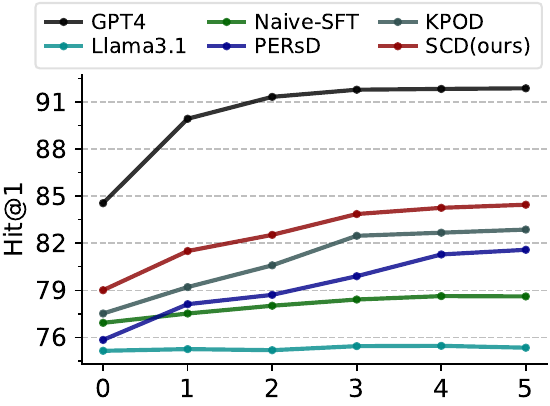}
    \label{fig:hyper_mct}}
    \caption{Hyperparameter analysis on WebQSP.}
    \label{fig:hyper}
\end{figure}
Figure \ref{fig:hyper} shows that increasing the number of error correction demonstrations and correction times slightly improves the performance of large-scale LLMs and small-scale LLMs fine-tuned with PERsD, KPOD, and SCD. However, it has little effect on the original Llama3.1 or models fine-tuned with Naive-SFT, which lack special training for error correction. 
Figure \ref{fig:hyper_mct} also highlights that large-scale LLMs have stronger error correction abilities than small-scale LLMs: GPT4 requires a maximum of 2 rounds of error correction to correct most errors, while small-scale LLMs of PERsD, KPOD, and SCD need 3 rounds or more. Balancing cost and performance, we set the final correction demonstration number to 8 and the maximum correction times to 3 for all experiments.

\subsection{QA on Unseen Dataset}
\input{table/unseen_dataset}
We evaluate small-scale LLMs on the unseen QA dataset TabFact~\cite{DBLP:conf/iclr/ChenWCZWLZW20} and show results in Table \ref{tab:unseen_datasets}.
Our small-scale LLM, significantly smaller than StructGPT (a unified structured data QA framework based on GPT3.5), achieves comparable results, demonstrating our approach's strong generalization. 
Compared to other knowledge distillation methods, our small-scale LLM exhibits superior generalization on new data. 

\section{Limitations}
\label{sec:limitation}
We observe four failure patterns where small-scale LLM fails to produce correct answers: (1) Correction threshold exceeded: The maximum correction iterations are reached, typically for questions that are too difficult for current model.
(2) False negative corrections: Valid queries returning ``null/none'' values are misidentified as ``empty mid-step output'' execution error, causing EPM to erroneously modify correct queries.
(3) Base model capability gap: 
For queries requiring ``LLM function'' of TrustUQA~\cite{DBLP:conf/aaai/ZhangJZ0HWHLC25}, the 8B student model underperforms GPT4 due to inherent capacity limitations. Based on the same query, large-scale LLMs can output the correct answer but small-scale LLMs cannot. 
(4) Alignment errors: Queries are successfully parsed and executed yet yield incorrect answers, primarily due to relation or entity mapping errors during the parameter-value alignment step \cite{DBLP:conf/aaai/ZhangJZ0HWHLC25}.

\section{Conclusion}
In this work, we propose Self-Correction Distillation (SCD), a novel CoT-distillation method for enhancing structured data QA capabilities in small-scale LLM. SCD integrates two key components: (1) an error prompt mechanism (EPM) that provides detailed error location and correction guidance during inference, and (2) a two-stage distillation strategy that effectively transfers both query-generation and error-correction competencies from large-scale LLM to small-scale LLM.
Experimental results demonstrate  that SCD achieves state-of-the-art performance on 8B parameter LLM with superior generalization capabilities. And GPT4 equipped with EPM surpasses SOTA benchmarks across most datasets. In the future, we focus on addressing the limitations outlined in Section \ref{sec:limitation}, including false negative cases where ``null/none'' values trigger execution error, and relation and entity alignment errors.

\section*{Acknowledgments}
This work is founded by National Natural Science Foundation of China (NSFC62306276/NSFCU23B2055), Yongjiang Talent Introduction Programme (2022A-238-G), and Fundamental Research Funds for the Central Universities (226-2023-00138). This work was supported by Ant Group.

\bibliography{aaai2026}
\input{appendix}
\end{document}

%% file: table/error_messages.tex
\begin{table*}[!hbpt]
    \centering
    \setlength{\tabcolsep}{4pt} 
    \resizebox{1.0\textwidth}{!}{
    \begin{tabular}{l|l|l|l}
    \toprule
                              & \textbf{Error Type}                             & \textbf{Example}                                                               & \textbf{Error Message}                                                                                          \\
                              \midrule
\multirow{11}{*}{\textbf{PE}} & \multirow{2}{*}{Undefined function name}        & \multirow{2}{*}{\textit{subtract(set1=..., set2=...)}}                         & \textit{The function `subtract' is not defined! Please call one of: [`get\_information', `min', `mean',}        \\ 
                              &                                                 &                                                                                & \textit{`max', `count', `sum', `keep', `set\_intersection', `set\_union', `set\_negation', `set\_difference'...} \\ \cline{2-4} 
                              & Illegal parameter                               & \textit{max(set=..., key=...)}                                                 & \textit{For function `max', parameter name `key' is illegal, the parameter name must be in [`set'].}            \\ \cline{2-4} 
                              & \multirow{2}{*}{Inconsistent parameters}        & \multirow{2}{*}{\textit{gi(head\_entity=..., relation=..., tail\_entity=...)}} & \textit{For function `get\_information', it is not allowed to assign values to parameters [`head\_entity',}     \\
                              &                                                 &                                                                                & \textit{`relation', `tail\_entity'] at the same time.}                                                          \\ \cline{2-4} 
                              & \multirow{2}{*}{Illegal comparator}             & \multirow{2}{*}{\textit{gi(tail\_entity=..., relation=..., head\_entity$<$...)}} & \textit{In function `get\_information', comparison symbol `$<$' for `head\_entity' is illegal, and}     \\
                              &                                                 &                                                                                & \textit{non-equal comparators are only allowed for parameters `tail\_entity' and `value'.}                                \\ \cline{2-4} 
                              & \multirow{2}{*}{Non-atomic operation}           & \multirow{2}{*}{\textit{sum(set=set\_negation(...))}}                          & \textit{The query is not an atomic operation: functions `sum' and `set\_negation' are nested.}                  \\ 
                              &                                                 &                                                                                & \textit{Please make sure that each step is atomic.}                                                             \\ \cline{2-4} 
                              & \multirow{2}{*}{Other non-standard expressions} & \multirow{2}{*}{\textit{sum(set=[output\_of\_query1, output\_of\_query2])}}    & \textit{Parsing the passed parameter value `[output\_of\_query1, output\_of\_query2]' failed.}                  \\
                              &                                                 &                                                                                & \textit{Please ensure that the format of the query is correct}                                                  \\ \hline
\multirow{5}{*}{\textbf{EE}}  & \multirow{2}{*}{Exception from Python executor} & \textit{sum(set=output\_of\_query1)}                                           & \textit{Exception from Python in function `sum': unsupported operand type(s) for +: `int' and `str'}            \\
                              &                                                 & // elements of output\_of\_query1 should be numeric                            & // a string-type element in output\_of\_query1                                                                  \\ \cline{2-4} 
                              & \multirow{3}{*}{Empty mid-step result}          & \textit{gi(relation=`Hometown', tail\_entity=`Utah')}                          & \textit{For query1, the execution result=set(), that is output\_of\_query1 is empty, which may affect}          \\
                              &                                                 & // 1st-step query to find the entity from Utah                                & \textit{subsequent query execution and final result. Please verify the correctness of entity or relation.}                   \\
                             &            &  // Data has 2 ``from'' relations: Hometown, Colleges    &     //  Incorrect relation led to empty outcome. The correct relation is ``Colleges''.                                             \\
                              \bottomrule
\end{tabular}
}
    \caption{Error Messages. PE and EE denote parsing error and execution error. \textit{gi} denotes \textit{get\_information}.}
    \label{tab:error_msg}
\end{table*}

%% file: table/datasets.tex
\begin{table}[ht]
\setlength{\tabcolsep}{2pt} 
    \centering
    \resizebox{0.49\textwidth}{!}{
    \begin{tabular}{lcclccclc}
    \toprule
\multirow{2}{*}{\textbf{Datasets}} & \multicolumn{2}{c}{\textbf{\# SFT}} &  & \multicolumn{3}{c}{\textbf{\# Raw}}           &  & \textbf{Other Information}  \\ \cline{2-3} \cline{5-7}
                                   & \textit{train}    & \textit{dev}    &  & \textit{train} & \textit{dev} & \textit{test} &  & \textit{train/dev/test}     \\
\midrule
\textbf{WTQ}                       & 11,322            & 2,830           &  & 14,152         & -            & 4,344         &  & 1,679/-/421 tables          \\
\textbf{WikiSQL}                   & 56,351            & 3,000           &  & 56,351         & 8,421        & 15,878        &  & 18,585/2,716/5,230 tables   \\
\textbf{MetaQA}                    & 30,000            & 6,000           &  & 329,282        & 39,138       & 39,093        &  & 134,741 triples             \\
- 1-hop                            & 9,000             & 2,000           &  & 96,106         & 9,992        & 9,947         &  & simple 1 hop                \\
- 2-hop                            & 10,500            & 2,000           &  & 118,980        & 14,872       & 14,872        &  & 21 types                    \\
- 3-hop                            & 10,500            & 2,000           &  & 114,196        & 14,274       & 14,274        &  & 15 types                    \\
\textbf{WebQSP}                    & 2,788             & 308             &  & 2,788          & 308          & 1,639         &  & retrieved version           \\
\textbf{CronQuestion}              & 25,000            & 4,000           &  & 350,000        & 30,000       & 30,000        &  & 323,635/5,000/5,000 triples \\
- Simple                           & 13,000            & 2,800           &  & 181,302        & 21,338       & 17,142        &  & simple time and entity      \\
- Complex                          & 12,000            & 1,200           &  & 168,698        & 8,662        & 12,858        &  & 3 complex types             \\
- Entity                           & 16,204            & 2,485           &  & 225,672        & 19,362       & 19,524        &  & query for entity            \\
- Time                             & 8,796             & 1,515           &  & 124,328        & 10,638       & 10,476        &  & query for time             
 \\ \bottomrule            
\end{tabular}
    }
    \caption{Statistics of experimental datasets.}
    \label{tab:datasets}
\end{table}

%% file: table/main_result.tex
\begin{table*}[!hbpt]
    \centering
    \resizebox{1.\textwidth}{!}{
    \begin{tabular}{lcclcclcccccclcclcccccccccc}
    \toprule
\multirow{2}{*}{Method} & \multicolumn{5}{c}{\textbf{Table QA}}                                                      &  & \multicolumn{9}{c}{\textbf{KG QA}}                                                                                                                              &  & \multicolumn{10}{c}{\textbf{Temporal KG QA}}                                                                                                                                               \\ \cline{2-6} \cline{8-16} \cline{18-27} 
                        & \multicolumn{2}{c}{\multirow{2}{*}{WikiSQL}} &  & \multicolumn{2}{c}{\multirow{2}{*}{WTQ}} &  & \multicolumn{6}{c}{MetaQA}                                                                                     &  & \multicolumn{2}{c}{\multirow{2}{*}{WebQSP}} &  & \multicolumn{10}{c}{CronQuestion}                                                                                                                                                          \\ \cline{8-13} \cline{18-27} 
                        & \multicolumn{2}{c}{}                         &  & \multicolumn{2}{c}{}                     &  & \multicolumn{2}{c}{\textit{1-hop}} & \multicolumn{2}{c}{\textit{2-hops}} & \multicolumn{2}{c}{\textit{3-hops}} &  & \multicolumn{2}{c}{}                        &  & \multicolumn{2}{c}{\textit{All}}   & \multicolumn{2}{c}{\textit{Complex}} & \multicolumn{2}{c}{\textit{Simple}} & \multicolumn{2}{c}{\textit{Entity}} & \multicolumn{2}{c}{\textit{Time}}  \\
                        \midrule
SOTA perf.              & \multicolumn{2}{c}{89.5$^\dagger$}             &  & \multicolumn{2}{c}{66.7$^\ast$}             &  & \multicolumn{2}{c}{98.4$^\ast$}    & \multicolumn{2}{c}{99.9$^\ast$}     & \multicolumn{2}{c}{99.4$^\ast$}     &  & \multicolumn{2}{c}{85.7$^\diamond$}               &  & \multicolumn{2}{c}{97.2$^\ddagger$} & \multicolumn{2}{c}{95.4$^\ddagger$}   & \multicolumn{2}{c}{99.5$^\ddagger$}  & \multicolumn{2}{c}{96.1$^\ddagger$}  & \multicolumn{2}{c}{99.1$^\ddagger$} \\
\hline
                        & \textit{w/}          & \textit{w/o}         &  & \textit{w/}          & \textit{w/o}         &  & \textit{w/}      & \textit{w/o}    & \textit{w/}      & \textit{w/o}     & \textit{w/}      & \textit{w/o}     &  & \textit{w/}          & \textit{w/o}         &  & \textit{w/}      & \textit{w/o}    & \textit{w/}       & \textit{w/o}     & \textit{w/}      & \textit{w/o}     & \textit{w/}      & \textit{w/o}     & \textit{w/}      & \textit{w/o}    \\
teacher(GPT4)           & \textbf{91.1}        & 87.7                 &  & 53.2                 & 48.5                 &  & \textbf{98.6}    & 97.2            & 99.4             & 98.2             & 99.3    & 98.3             &  & \textbf{91.7}        & 84.4                 &  & \textbf{98.1}    & 97.6            & \textbf{97.3}     & 96.5             & \textbf{99.7}    & 99.6             & \textbf{96.9}    & 96.2             & \textbf{99.6}    & 99.5            \\ \hline
                        
Llama3.1                & 74.5                 & 73.4                 &  & 35.1                 & 33.9                 &  & 94.6             & 93.5            & 91.1             & 88.7             & 83.4             & 81.9             &  & 75.2                 & 74.9                 &  & 88.1             & 87.8            & 80.9              & 80.7             & 97.8             & 97.5             & 83.2             & 82.9             & 96.9             & 96.8            \\
Naive-SFT               & 79.3                 & 77.1                 &  & 42.2                 & 40.7                 &  & 96.1             & 94.5            & 93.6             & 92.7             & 88.1             & 86.2             &  & 78.2                 & 76.7                 &  & 92               & 90.8            & 86.4              & 85.3             & 99.2             & 99.1             & 88.6             & 87.2             & 98.7             & 96.5            \\
PERsD         & 84.2           & 79.6           &           & 44.3                & 40.4               &           & 97.2                & 96.2               & 98.1           & 94.8           & 97.3           & 93.2           &  & 79.7          & 75.6          &  & 93.4             & 91.2            & 89.1             & 86.1            & 99.2             & 99.1            & 91.4             & 87.7            & 97.2             & 96.5            \\
KPOD                    & 86.1                 & 81.8                 &  & 46.5                 & 42.7                 &  & 97.1             & 96.8            & \textbf{98.5}    & 96.7             & 97.8             & 95.6             &  & 82.3                 & 77.3                 &  & 95.5             & 92.1            & 92.4              & 87.9             & 99.3             & 98.6             & 94.2             & 88.7             & 98.4             & 97.1            \\
SCD(ours)               & \textbf{86.9}        & \textbf{83.1}        &  & \textbf{48.9}        & \textbf{43.9}        &  & \textbf{97.4}    & \textbf{97.1}   & 98.4             & \textbf{97.2}    & \textbf{98.2}    & \textbf{96.3}    &  & \textbf{83.7}        & \textbf{78.8}        &  & \textbf{96.9}    & \textbf{92.6}   & \textbf{95.1}     & \textbf{88.8}    & \textbf{99.5}    & \textbf{99.2}    & \textbf{95.8}    & \textbf{89.2}    & \textbf{98.8}    & \textbf{97.3}  \\ \bottomrule
\end{tabular}
}
    \caption{Denotation accuracy of TableQA, Hit@1 of KG QA and temporal KG QA. \textit{w/} denotes ``inference w/ EPM'', and \textit{w/o} denotes ``inference w/o EPM''. The SOTA performance is the best-known results from TAPEX$^\dagger$~\cite{DBLP:conf/iclr/LiuCGZLCL22}, Readi$^\ast$~\cite{DBLP:conf/acl/ChengZXYZQHCL0R24}, RoG$^\diamond$~\cite{DBLP:conf/iclr/LuoLHP24}, and TrustUQA(GPT3.5)$^\ddagger$~\cite{DBLP:conf/aaai/ZhangJZ0HWHLC25}. The bold results in teacher represent exceeding SOTA, and bold in students is the best of all students.}
    \label{tab:main_result}
\end{table*}

%% file: table/ablation_detail.tex
\begin{table*}[!hbpt]
    \centering
    \setlength{\tabcolsep}{6pt} 
    \resizebox{0.8\textwidth}{!}{
    \begin{tabular}{lcclcccclclccccc}
    \toprule
                             & \multirow{2}{*}{\textbf{WikiSQL}} & \multirow{2}{*}{\textbf{WTQ}} &  & \multicolumn{4}{c}{\textbf{MetaQA}}                                &  & \multirow{2}{*}{\textbf{WebQSP}} &  & \multicolumn{5}{c}{\textbf{Cronquestion}}                                     \\ \cline{5-8} \cline{12-16} 
                             &                                   &                               &  & \textit{avg.} & \textit{1-hop} & \textit{2-hops} & \textit{3-hops} &  &                                  &  & \textit{All}  & \textit{Com}  & \textit{Sim}  & \textit{Ent}  & \textit{Tim}  \\
\midrule
SCD                          & \textbf{86.9}        & \textbf{48.9}        &  & \textbf{98.0}   & \textbf{97.4}  & \textbf{98.4}   & \textbf{98.2}   &  & \textbf{83.7}        &  & \textbf{96.9} & \textbf{95.1} & \textbf{99.5} & \textbf{95.8} & \textbf{98.8} \\
\textit{- self-distillation} & 86.1                 & 46.2                 &  & 97.1          & 97.1           & 97.2            & 97.1            &  & 82.2                 &  & 95.3          & 93.6          & 99.2          & 92.6          & 98.5          \\
\textit{- error-msg in EPM}  & 83.7                 & 44.8                 &  & 97.2          & 97.2           & 97.7            & 96.8            &  & 79.5                 &  & 93.8          & 90.9          & 99.1          & 90.9          & 97.7         \\ \bottomrule
\end{tabular}
}
    \caption{Ablation Study. \textit{- self-distillation} means to train student with only teacher-distillation stage. \textit{- error-msg in EPM} means EPM only signals error occurrence but does not provide details on error type and specific messages.}
    \label{tab:ablation}
\end{table*}

%% file: table/unseen_dataset.tex
\begin{table}[!hbpt]
\setlength{\tabcolsep}{8pt} 
    \centering
    \resizebox{0.49\textwidth}{!}{
    \begin{tabular}{lcc}
    \toprule
          SOTA perf. of unified QA models  & \multicolumn{2}{c}{87.6$^\star$}       \\
          \hline
          & \textit{inference w/ EPM} & \textit{inference w/o EPM} \\
Llama3.1  &      64.0       &     60.8         \\
Naive-SFT &      75.4       &      69.3         \\
PERsD     &     79.8        &     74.7         \\
KPOD      &     79.2        & 
75.1         \\
SCD(ours) &     83.4        &  78.8         \\ \bottomrule
\end{tabular}
    }
    \caption{Accuracy on TabFact. SOTA performance of unified models is from StructGPT(GPT3.5)$^\star$~\cite{DBLP:conf/emnlp/JiangZDYZW23}.
    }
    \label{tab:unseen_datasets}
\end{table}

%% file: appendix.tex
\clearpage

\appendix
\section{Prompt Template}
\label{sec:appendix_prompt}
\begin{figure}[!htpb]
    \centering
    \subfigure[P\_Q: query generation prompt template]
    {\includegraphics[width=1.0\linewidth]{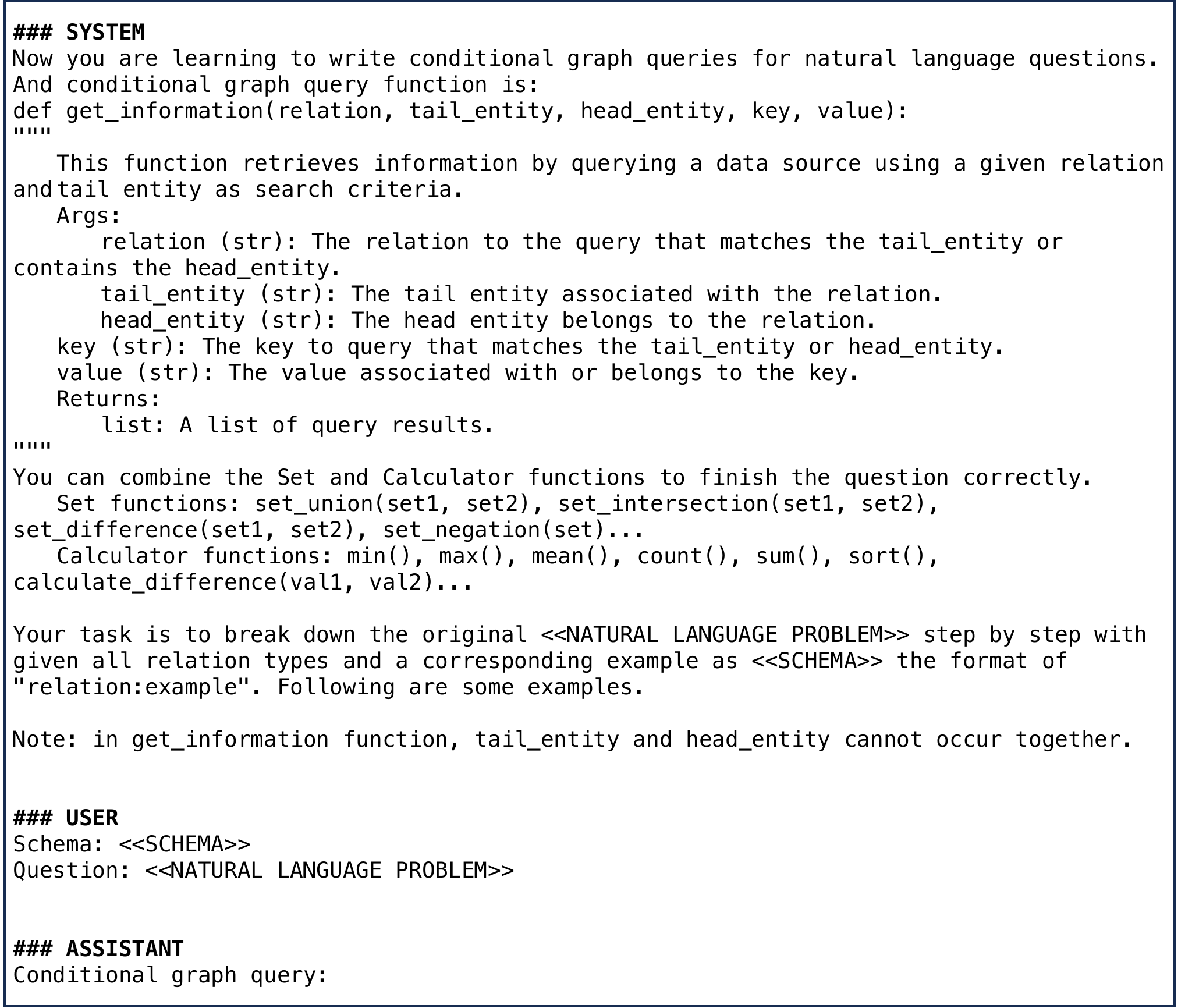}
    \label{fig:prompt-q}}
    \subfigure[P\_C: error correction prompt template]{\includegraphics[width=1.0\linewidth]{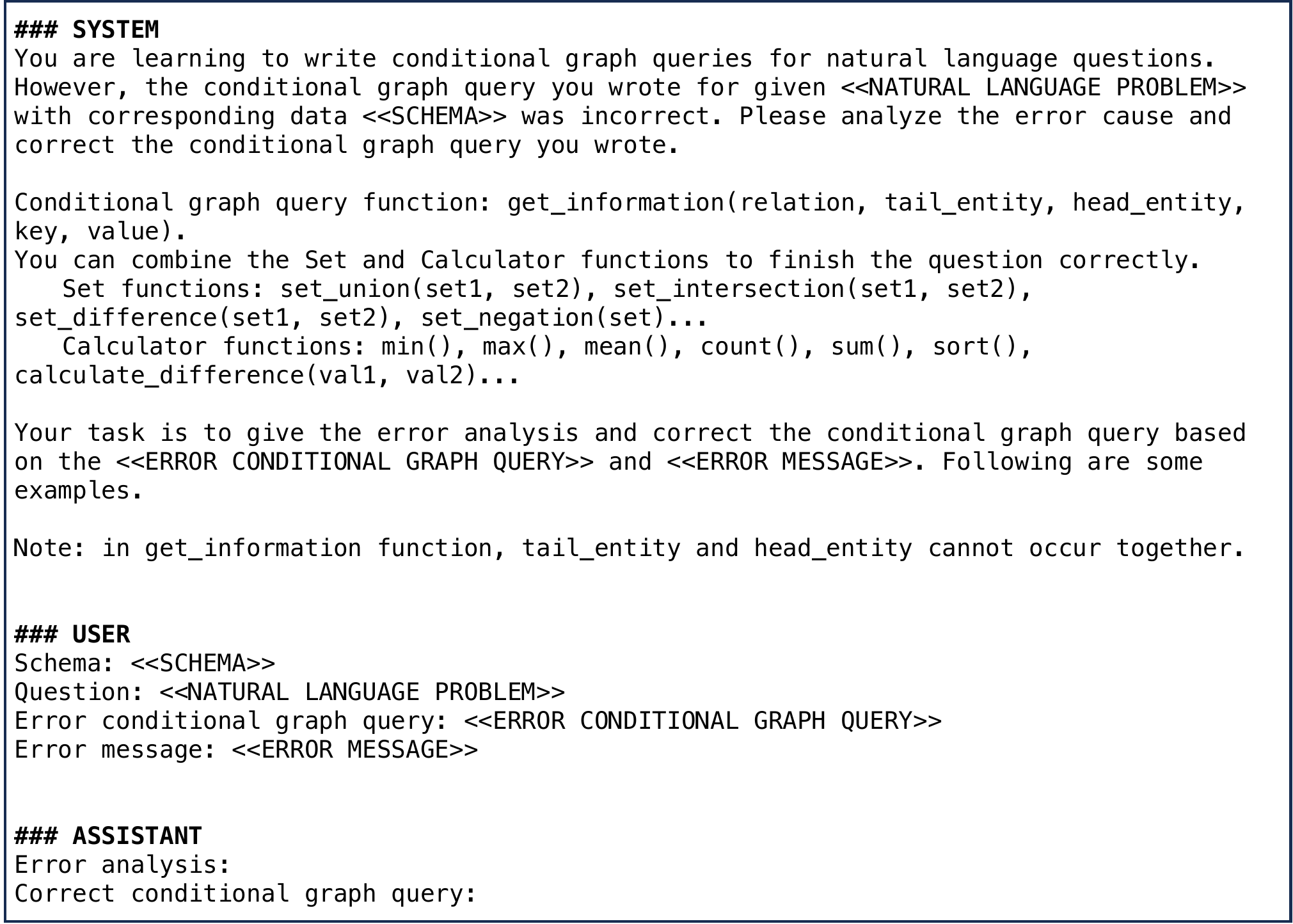}
    \label{fig:prompt-c}}
    \caption{Prompt templates to query generation and error correction. The prompts follow the Chat Completions format,  including roles of ``system'', ``user'', and ``assistant''. We prefix each role with \textit{\#\#\#} to distinguish it from the content.}
    \label{fig:prompt_template}
\end{figure}

Figure \ref{fig:prompt_template} shows our prompt template for query generation and error correction. And we keep the query generation prompt template the same as in TrustUQA~\cite{DBLP:conf/aaai/ZhangJZ0HWHLC25}, as detailed in Figure \ref{fig:prompt-q}. 

\section{More Details of EPM}
\label{sec:appendix_epm_details}
\subsection{Error Type Provenance}
For the sources of error classification, we used TrustUQA \cite{DBLP:conf/aaai/ZhangJZ0HWHLC25} framework to conduct pre-experiments, and manually analyzed the experimental results to summarize the most common types of errors. This method ensures that the error classification is practical and representative.

\subsection{Implementation Specification}
For the parser implementation, we refer to the TrustUQA \cite{DBLP:conf/aaai/ZhangJZ0HWHLC25} framework and use regular expressions to extract parameters and function names from queries. These extracted elements are compared against predefined function call syntax criteria to check whether parameter names, function names, and so on conform to the specification.

For parsing quality impact, since our error recognition is based entirely on the exact matching of regular expression extraction and predefined syntax, the current implementation achieves 100\% accuracy.
